\documentclass{bmcart}
\usepackage{amssymb,amsfonts,amsmath}
\usepackage{graphicx}
\usepackage{subcaption}
\usepackage{float}
\usepackage{parskip}
\usepackage{adjustbox}

\usepackage[plainpages=false,pdfpagelabels,pagebackref=false,naturalnames=true,hyperindex=true,pdftitle={Language statistics at different spatial, temporal, and grammatical scales}]{hyperref}
\usepackage[draft,inline,nomargin]{fixme} \fxsetup{theme=color}

\FXRegisterAuthor{cp}{acp}{\color{blue}CP}
\FXRegisterAuthor{cg}{cgg}{\color{red}CG}
\FXRegisterAuthor{am}{ajm}{\color{magenta}AM}
\FXRegisterAuthor{rl}{rla}{\color{green}RL}

\newcommand{\rme}{\mathrm{e}}

\begin{document}

\begin{frontmatter}

\begin{fmbox}
\dochead{Research}

\title{Language statistics at different spatial, temporal, and grammatical scales}

\author[
   addressref={Ciencias},
   email={fernanda@ciencias.unam.mx}
]
{\inits{FSP}\fnm{Fernanda} \snm{Sánchez-Puig}}
\author[
   addressref={Ciencias},
   email={rogelio97@ciencias.unam.mx}
]{\inits{RLZ}\fnm{Rogelio} \snm{Lozano-Aranda}}
\author[
   addressref={Posgrado},
   email={dante.perez.m@gmail.com}
]{\inits{DPM}\fnm{Dante} \snm{Pérez-Méndez}}
\author[
   addressref={Roslin},
   email={ecolman@ed.ac.uk}
]{\inits{EC}\fnm{Ewan} \snm{Colman}}

\author[
   addressref={MIT},
   email={alfredom@mit.edu}
]{\inits{AJ}\fnm{Alfredo J.} \snm{Morales-Guzmán}}
\author[
   addressref={fisica},
   email={carlosp@fisica.unam.mx}
]{\inits{CP}\fnm{Carlos} \snm{Pineda}}
\author[
   addressref={c3},                  
   email={pedro.rivera@c3.unam.mx}
]{\inits{PR}\fnm{Pedro Juan} \snm{Rivera Torres}}
\author[
   addressref={iimas, c3, lakeside},                  
   corref={iimas},               
   email={cgg@unam.mx}
]{\inits{CG}\fnm{Carlos} \snm{Gershenson}}

\address[id=Posgrado]{
\orgname{Posgrado en Ciencias de la Computación},
\street{Universidad Nacional Autónoma de México}, 
\city{Mexico City}, 
\postcode{04510},
\cny{Mexico}
}
\address[id=Roslin]{
\orgname{Roslin Institute},
\street{University of Edinburgh}, 
\city{Midlothian}, 
\cny{United Kingdom}
}
\address[id=MIT]{
\orgname{MIT Media Lab},
\street{Cambridge}, 
\city{MA}, 
\postcode{02139}, 
\cny{USA}
}
\address[id=fisica]{
\orgname{Instituto de Física},
\street{Universidad Nacional Autónoma de México}, 
\city{Mexico City}, 
\postcode{04510}, 
\cny{Mexico}
}
\address[id=iimas]{
\orgname{Instituto de Investigaciones en Matemáticas Aplicadas y Sistemas},
\street{Universidad Nacional Autónoma de México}, 
\city{Mexico City}, 
\postcode{04510}, 
\cny{Mexico}
}
\address[id=c3]{
\orgname{Centro de Ciencias de la Complejidad},
\street{Universidad Nacional Autónoma de México}, 
\city{Mexico City}, 
\postcode{04510}, 
\cny{Mexico}
}
\address[id=lakeside]{
\orgname{Lakeside Labs GmbH},
\street{Lakeside Park B04}, 
\city{Klagenfurt am Wörthersee}, 
\cny{Austria}
}
\address[id=Ciencias]{
\orgname{Facultad de Ciencias},
\street{Universidad Nacional Autónoma de México}, 
\city{Mexico City}, 
\postcode{04510}, 
\cny{Mexico}
}

\end{fmbox}

\begin{abstractbox}
\begin{abstract}
Statistical linguistics has advanced considerably in recent decades as data has
become available. This has allowed researchers to study how statistical
properties of languages change over time. In this work, we use data from
Twitter to explore English and Spanish considering the \emph{rank diversity}
at different scales: temporal (from 3 to 96 hour intervals), spatial (from 3km
to 3000+km radii), and grammatical (from monograms to pentagrams). We find that
all three scales are relevant. However, the greatest changes come from
variations in the grammatical scale. At the lowest grammatical scale
(monograms), the rank diversity curves are most similar, independently on the
values of other scales, languages, and countries. As the grammatical scale
grows, the rank diversity curves vary more depending on the temporal and spatial
scales, as well as on the language and country. We also study the statistics of
Twitter-specific tokens: emojis, hashtags, and user mentions. These particular type of tokens show a sigmoid kind of behaviour as a rank diversity function.  Our results are
helpful to quantify aspects of language statistics that seem universal and what
may lead to variations.
\end{abstract}


\end{abstractbox}
\end{frontmatter}

\section{Introduction} 

Statistical linguistics have become a relevant field of research
over the last century~\cite{zipf}.  In this context, random-text models have
been proposed as an explanation for the so-called Zipf's law~\cite{Booth1967386,Montemurro2001567,Cancho2002,newman2005power,1367-2630-13-4-043004,PhysRevE.83.036115}.
Random texts and real texts are compared showing that real texts fill the
lexical spectrum much more efficiently and regardless of the word length,
suggesting that the meaningfulness of Zipf's law is high.  Other studies have
focussed on language origins and evolution~\cite{Gell-Mann18102011,Kirby20032007,Steels2012,Baronchelli,SOLE201647}. The recent
availability of data enables further analysis of language usage including
dynamics and changes over time~\cite{Michel14012011,Perc07122012,PhysRevX.3.021006,Cocho2015,Alshaabieabe6534}.  

Previous studies consider language variation in timescales of years and
centuries~\cite{Michel14012011,Perc07122012}. In this
article, we study the use and change of language use 
at the timescales of
hours and days. 
We use geolocalized Twitter data to compare languages at
different timescales, as well as at different
spatial scales and ``grammatical scales''.
Our motivation was to measure the relevance of each scale in language statistics. In other words, does language use vary more with time, space, or structure?


Twitter data has been used for studying language in the context of sentiment and topic analysis, (mis)information spreading, and activity patterns~\cite{Bollen:2011,Dodds2011,Morales2014,doi:10.1098/rsif.2016.1048,Alshaabieabe6534,Pennycook2021}. The information in the meta-data, includes location, time and text, which enables the analysis of dynamics and geography at multiple scales. Previous studies have found differences in the way people use text and interaction mechanisms such as URLs, hashtags, mentions, replies, and retweets by country or culture~\cite{Hong2011,Weerkamp2011}. Moreover, the usage of Twitter hashtags showed consistencies with the distribution of wealth in urban areas~\cite{Morales2019}. 

Our previous research shows that changes in word usage
within certain languages follow the same pattern.
We measured these changes using a metric we define as
\emph{rank diversity}. To calculate it, we consider a corpus of words ranked by their frequency (number of times it appears on a given time interval), and then counting how many different words occupied each rank (see Methods).
If a rank is occupied by a single word at all times, the rank diversity is minimum. On the other hand, if
in each time interval we have a different word on a given rank, then its rank
diversity is maximum. If we plot rank diversity as a function
of the rank, we can analyze how word usage changes in time.
In~\cite{Cocho2015,10.3389/fphy.2018.00045}, we compared the rank diversity of
books in different languages. It was shown that rank diversity as a function of
the rank can be approximated with a sigmoid, with similar parameters for all languages studied.

Twitter data has unique characteristics that make it an interesting studying subject. 
Unlike books or other written pieces, people can only publish tweets with a
limited number of characters, making interesting the study of use of language in a medium that allows only very short texts and whether it
differs statistically from longer texts.
Also, Twitter offers a much finer temporal dimensionality than
physically published material. And since tweets can be geotagged, we can study geographical differences of language use at very fine scales. Moreover, due to its social network nature,
interactions between users (mentions, retweets) and trending topics create a
unique language ecosystem. Finally, it provides a big dataset, suitable to perform statistical 
analysis.

In this work, we analyzed more than 20 million geolocalized
tweets from eight different countries.  We calculated rank diversity in different spatial, temporal, and
grammatical scales.
We were interested in measuring the changes in rank diversity for different
scales considered. We
observed several features. First, higher scales are related to higher
rank diversities except in the case of time, which exhibits a concave behavior,
where shorter and larger time intervals have higher rank diversity than medium
ones. Second, different types of scale affect each other, \emph{i.e.}, they are not independent.
Finally,
considering the importance of a scale as the rank diversity average dispersion
in that scale, we found that the grammatical scale is the most important among the
three scales. Temporal and spatial scales have similar importance in the
Spanish-speaking countries, while the spatial scale is the least important in
English-speaking countries.


\section{Methods and Data} 

We define rank diversity $d(k)$ as the number of words occupying a given rank
$k$ during the period of time of the study divided by the number of time
intervals $T$. Therefore, rank diversity is given by:
\begin{equation}
d(k)=\frac{|X(k)|}{T},
\end{equation}
where $|X(k)|$ is the cardinality (\emph{i.e.}, number of unique words) that appear at
rank $k$ during all time intervals.
The time between time ``slices'' is
$\Delta t$, so that the total time considered is $T\Delta t$.

We have found that rank diversity curves for six different Indoeuropean
languages are very similar, as they can be fitted with a sigmoid curve with
small differences between
languages~\cite{Cocho2015,10.3389/fphy.2018.00045,Cocho2019}. This pattern is
also present in the rank dynamics of sports~\cite{Morales2016} and other
systems~\cite{Iniguez2022}.

We use geo-located Twitter data to analyze changes in language
usage. The tweets were collected
using the Twitter Streaming Application Programming Interface (API). We
consider over 20 million geo-located tweets posted from Argentina, Canada,
Colombia, India, Mexico, South Africa, Spain and the United Kingdom during 2014
(when each tweet was limited to 140 characters and threads were not as commonly
used) and we calculate rank diversity using different time intervals $\Delta
t$. Geo-located tweets contain precise latitude and longitude coordinates at
the moment of their creation. Twitter activity has been previously analyzed to
understand patterns of global synchronization
\cite{doi:10.1098/rsif.2016.1048}, spreading mechanisms~\cite{Morales2014} and
political polarization~\cite{doi:10.1063/1.4913758}.

We define the ``grammatical scale'' as the length of $N$-gram blocks used ($N=1,2,3,4,$ and $5$)~\cite{Michel14012011}.
Single words are monograms or 1-grams, sets of five words
are pentagrams or 5-grams, etc.  We have previously studied how the grammatical scale
affects the rank dynamics of words using the Google Books $N$-gram
dataset~\cite{10.3389/fphy.2018.00045}. We found that the grammatical scale
varies language statistics (rank diversity, change probability, rank entropy
and rank complexity) more than changes of the language.
In other words, a change in the 
grammatical scale implies a greater change in the statistics than a change of
language (among English, Spanish, French, German, Italian, and Russian).

To define the temporal scale, we need to remember that we define rank diversity
$d(k)$ as the number of words occupying a given rank $k$ across all times,
divided by the number of time intervals $T$ considered. We can change the time
interval $\Delta t$ and calculate rank diversity for different values of  $T$.
It should be noted that if the same dataset is used, as the temporal scale
$\Delta t$ increases, $T$ will decrease.  

\begin{figure}
    \centering
    \includegraphics[scale=0.23]{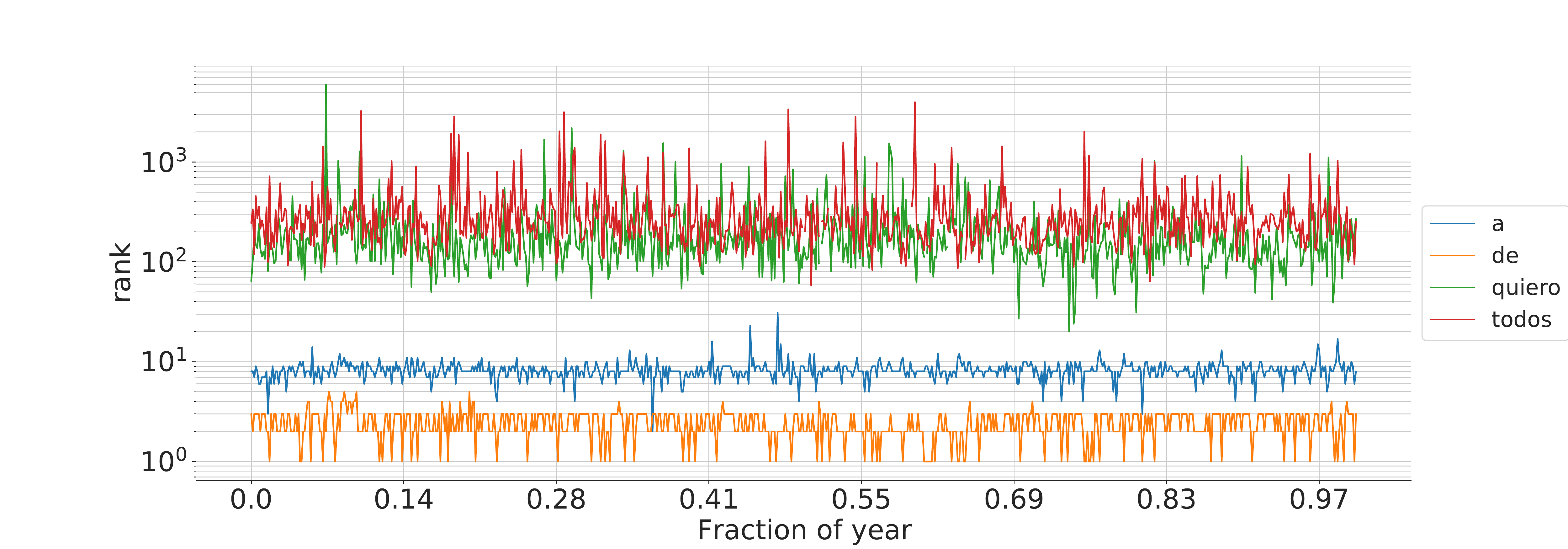}
    \captionsetup{width=.92\linewidth}
    \caption{Rank evolution of some Spanish words during 2014. Note that if a  word has a high rank, its trajectory takes a wider set of possible ranks compared  to  words with lower ranks~\cite{Cocho2015}.
}
    \label{fig:rank_evol}
\end{figure}

To illustrate the rank evolution of words (1-grams, but the same applies to any
grammatical scale), 
we  show examples of some arbitrarily chosen
Spanish words in Figure~\ref{fig:rank_evol}. For example, in this case $ d (k = 1) $ is obtained by dividing
the number of unique series, which represent words graphically, that at some time passes through the line representing point $ k = 1 $, over the total number
of 3 hourly intervals that divide a year. 

We study the effects of the spatial scale tweets from Mexico, Spain, Argentina,
and the United Kingdom only, since these countries are the only ones with enough geolocalized data to be statistically significant at different spatial scales (about 3.9, 3.7, 4.6, and
5.6 million tweets from Mexico, Spain, Argentina, and the United Kingdom,
respectively). For the first scale, we make a circle with a 3km radius located
in the geographical center of the capital city (Mexico City, Madrid, Buenos
Aires, and London). For the next scales, we increment the radius of the circle
by a power of two each time, \emph{i.e.}, 6km, 12km,24km, 48km, 98km,..., until
we include the whole country. To avoid biased results, we kept the same number
of tweets used in the analysis of the smallest spatial scale for the remaining
spatial scales for each country. For example, the number of tweets inside the
circle of radius of 3 km for Mexico was 309,792. Therefore, for each of the
remaining  spatial scales, which in this case are represented as the area
inside circles of increasing radius, we took a random sample of tweets of size
309,792 without replacement.

As there are three considered scales, to see any potential differences in the
behavior of rank diversity across values of these scales, we generated $I \cdot
J \cdot S $ rank diversity curves for each country, where $I$,$J$, and $S$ are
the number of different values that one particular scale can adopt. So we
generated rank diversity curves corresponding to each different combination of
values of the considered scales. For example, in the case of Mexico, we have $5
\cdot 6 \cdot 10 = 300$ possible combinations. In order to have a numeric value
that quantitatively summarizes the behavior of a rank diversity curve
(measuring that behavior as how fast rank diversity increases as a function of
rank) and, in consequence, simplify the description of the system and reduce
the observed complexity thereof, we used estimations of $\mu$, which is a
parameter of the sigmoid curve that indicates the rank where rank diversity
curves reach $1/2$. The sigmoid is the cumulative of a Gaussian
distribution,
\emph{i.e.}  
\begin{equation}
\Phi_{\mu,\sigma}(\log_{10} k)=\frac{1}{\sigma\sqrt{2\pi}}
    \int_{-\infty}^{\log_{10} k} \rme^{-\frac{(y-\mu)^2}{2\sigma^2}} {\rm d}y, 
\label{eq:sigmoid}
\end{equation}
and is given as a function of $\log k$~\cite{Cocho2015}.

The key concept to measure the relevance of changes in scales over rank diversity behavior is to understand that a lower value of $\mu$ indicates a greater speed of rank diversity increment as a function of rank. 

It is also important to remember that the introduction of both the rank diversity measure and the use of $\mu$ as a way of measuring its changes throughout scales implies the use of collective or aggregate measurements, so care must be taken when trying to generate particular conclusions to avoid making incorrect assumptions (ecological fallacy).

To measure the relative importance of a scale in terms of how much a change
between two different values of this particular scale influences changes in the
behavior of rank diversity, we used the following average:
\begin{equation}
    \eta(s) = \frac{\sum_{i}^{I} \sum_{j}^{J}\sigma^s_{i,j}}{I \cdot J}, 
    \label{eq:ETA}
\end{equation}
where $\sigma^s_{i,j}$ corresponds to the standard deviation of estimated
values of $\mu$ associated to the scale $s$ given fixed $i$ and $j$ values of the
two remaining scales,\emph{ i.e.}, if
\begin{equation}
   \Bar{\mu}_{i,j} = \sum_s \frac{\mu^s_{i,j}}{N},
   \label{eq:mean_MU}
\end{equation}
 then
\begin{equation}
    \sigma^s_{i,j} = \sqrt{ \frac{\sum_{s} (\mu^s_{i,j} - \Bar{\mu}_{i,j})^2}{N-1} }.
    \label{eq:Dispersion}
    \end{equation}
In other words, $\eta(s)$ is intended to capture the average dispersion of
$\mu$ in the scale $s$. This permits to objectively compare different scales
and determine which one is the most important in terms of modifying the speed
of rank diversity increment between different values of that scale. In the
Results section we show how $\eta(s)$ effectively quantifies what can be
visually seen using graphs of $\mu$ versus values of scales.

Finally, to statistically support the graphically observed results, we used an
additive linear regression model to perform the $F$-test and assess that at least
one of the scales has a significant effect on $\mu$ and therefore in rank
diversity.

Moreover, the $t$-test of each coefficient associated to independent variables is
used to assess whether or not each individual scale is contributing to explain
the variability of $\mu$ assuming a linear model, or in the case of the
coefficients that represent multiplicative terms in the multiplicative model,
to assess if there is a statistically significant interaction between each pair
of scales. The effect of interactions between pairs of scales is shown by
observing that as one of the scales increases, the  behavior of rank diversity
depends on the specific value of the remaining scales. In this case,
interaction effects are subtle to observe graphically, so a statistical
approach is worth the effort to support the hypothesis that interactions exist.

These models were fitted using the $\log_{10}$ scaled values of temporal and
grammatical scales as predictors and $\mu$ as response. In particular, the
multiplicative model is 
\begin{equation}
    Y = \beta_0 + \beta_1  X_1 + \beta_2 X_2 + \beta_3 X_3 + \beta_4 X_1  X_2 
            + \beta_5 X_1  X_3 + \beta_6 X_2  X_3. 
\label{eq:multiplicative}
\end{equation}
This model is reduced to a linear one discarding the terms that contain
products of predictors ($X_i$). We can understand the coefficients ($\beta_i$) as
weights that determine the influence of an associated predictor (a particular
scale) over the response, which in this case is $\mu$. That is the reason why
we use hypothesis tests to evaluate whether there is evidence that a
coefficient is different from zero or not. In the products of two scales the
coefficients measure the effect of one scale over how the other influences the
response. We can note this by factoring a common linear predictor. For example
$X_1$: $Y = \beta_1 X_1 + \beta_4 X_1 X_2 + \beta_5 X_1 X_3 = (\beta_1 +\beta_4
X_2 + \beta_5 X_3) X_1  = \beta_{145}(X_2,X_3) X_1$, such that now
$\beta_{145}(X_2,X_3)$ act as a multivariate function slope of $X_1$ that
determines how $X_2$ and $X_3$ influences the relation of $X_1$ with $Y$.
Specifically, if the associated coefficients $\beta_4$ and $\beta_5$ of the
predictors $X_2$ and $X_3$ are statistically different from zero, then there is
evidence that this set of predictors interact with $X_1$ pairwise. $\beta_0$ geometrically corresponds to the level of the hyperplane that better
fits the observations. In this particular case, it does not give any further
information of interest. All our models were fitted via linear least-squares
problems solved by the QR factorization method (for numerical stability) using
the standard \texttt{lm} function of the programming language R.

\section{Results} 
\begin{figure}[h!]
\begin{center}
\includegraphics[width=1.1\linewidth,height=\textheight,keepaspectratio]{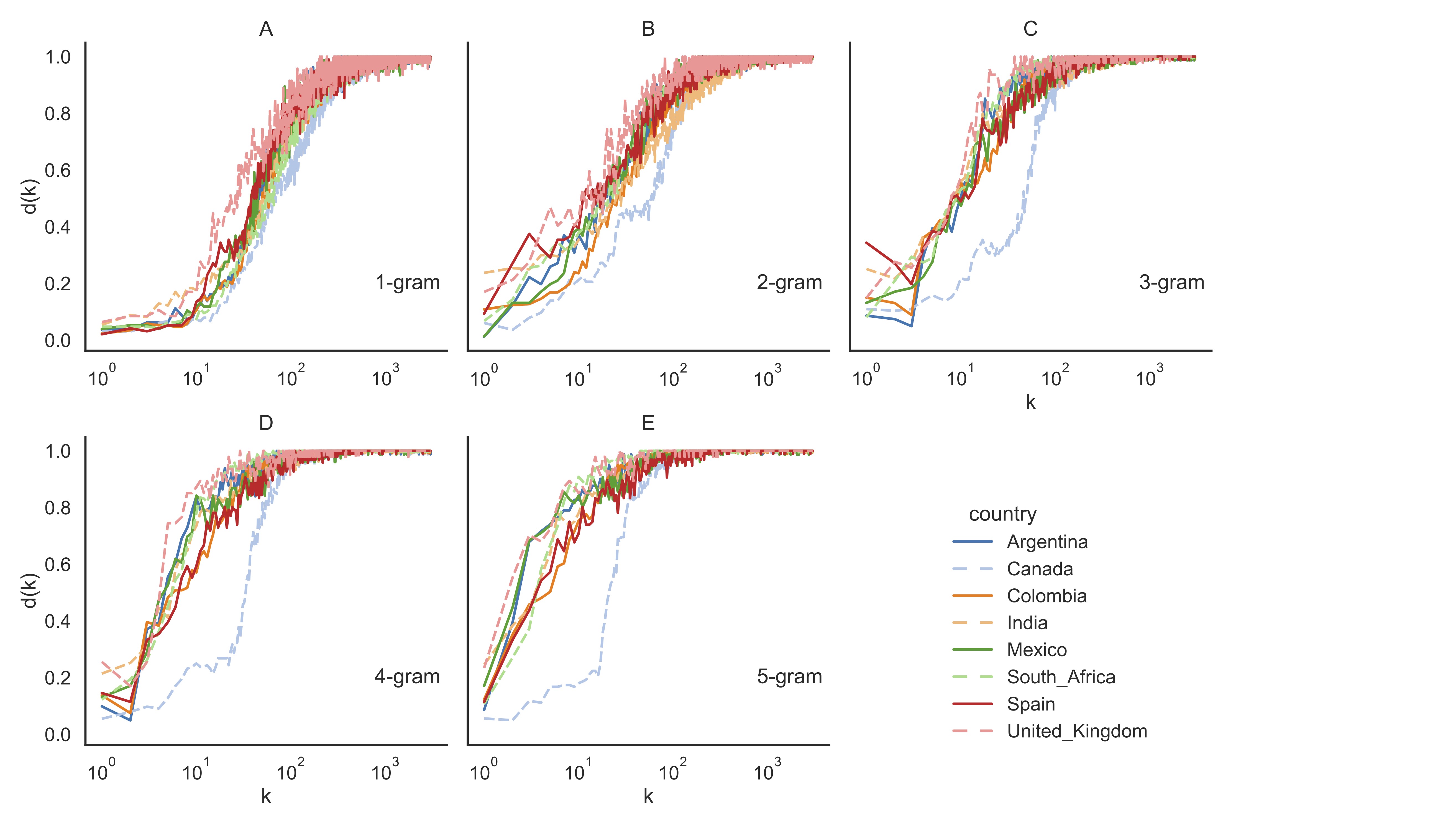}
\captionsetup{width=.92\linewidth}
\caption{Rank diversity $d(k)$ for eight different countries at different
grammatical scales $N=1,2,...,5$, $\Delta t = 24$hr,. Spanish-speaking
countries are shown with solid lines, English-speaking countries with dashed
lines. A) monograms, B) digrams, C) trigrams, D) tetragrams, and E) pentagrams.
X-axis is shown in log scale.}
\label{fig1}
\end{center}
\end{figure}

We calculate the rank diversity of $N$-grams for tweets from eight different countries: half
Spanish-speaking and half English-speaking. In Figure~\ref{fig1} , we plot the
rank diversity using a time interval $\Delta t$ of 24 hours for $N=1,2,...,5$.

First, we note that the sigmoid curve still provides an adequate description of
the rank diversity behavior when we consider small time intervals of
observation. This is also the case for all the considered combination of
scales. Also, we can observe that for $N=1$, the rank diversity fits are quite
similar. However, when we increment $N$, the fits separate from each other.
This difference suggests that 1-grams have a similar rank diversity
independently of language and country. However, for 2-grams, 3-grams, and
4-grams, we see a different pattern. The Spanish-speaking countries' curves are
close to each other, forming a cluster, whereas the United Kingdom and Canada
separate from the rest. This behavior means that there are grammatical and
geographical features that make them distinct from the rest. In the next
sections, we explore the effects at different scales (grammatical, spatial, and
temporal) on rank diversity using estimates of the parameter $\mu$ of the
sigmoid curves. 

\subsection{Grammatical Scale} 

Following the $N$ values in increasing order, Figures~\ref{fig:temp}
and~\ref{fig:geograph} show that as the grammatical scale increases, also the
speed of rank diversity increment increases. In general, this stands
independently of the country, whether the language is Spanish or English, or
which values the other two scales adopt. Note that a larger grammatical scale
means an increment in the complexity of the phrases. At the top of the scale
(5-grams), we have the rank diversity of blocks formed by five words. The
possible combinations of five-word blocks are larger than the ones of four
words, and these from the ones of three words. As a consequence, for the
initial ranks, we have more diversity than in lower grammatical scales.
Moreover, we can confirm that for 1-grams, $\mu$ is similar in both Spanish and
English and practically independent of the spatial scale. Namely, for words,
there are no changes in the speed of rank diversity increment for any area
analyzed. Nonetheless, it does vary with respect to the temporal scale as can be
seen in the first column of Figure~\ref{fig:temp}. This illustrates the
importance of using different scales to analyze the rank diversity of
languages.

Also, we can visually identify that changes in the grammatical scale account
for the greatest overall increase in speed of rank diversity increment compared
to the other two scales. At the end of the results section, we quantify this
qualitative observation comparing average dispersions.

\subsection{Temporal Scale} 

\begin{figure}
    \includegraphics[width=0.9\linewidth,height=\textheight,keepaspectratio]{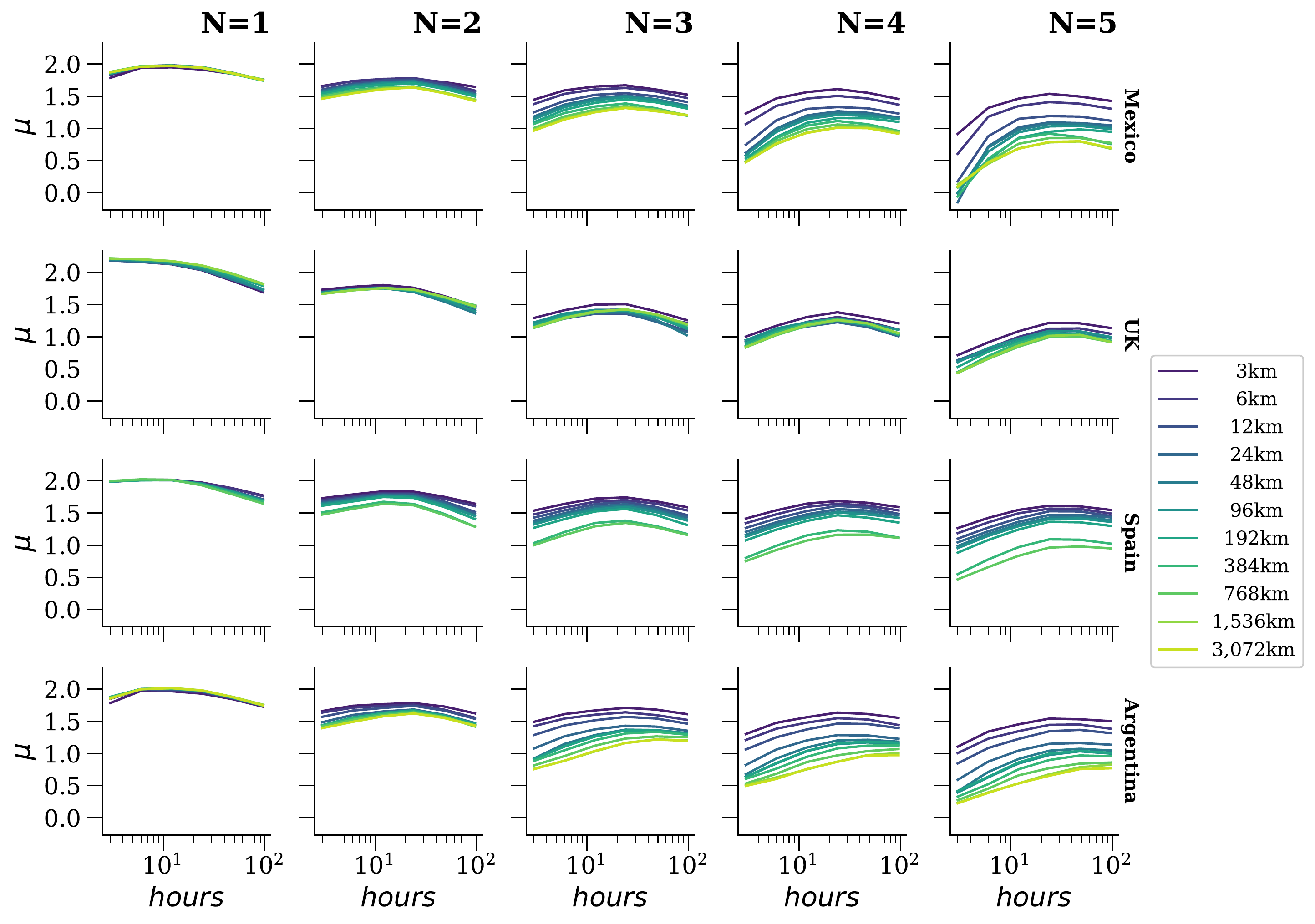}
    \captionsetup{width=.92\linewidth}
    \caption{Different estimations of $\mu$. Measure the speed of rank diversity increment vs. different values of the temporal scale. Note that the $x$-axis is
in $\log_{10}$ scale. The $N$ values represent $N$-grams (grammatical scale).
Each row uses Twitter data from the country specified at the right label.
Colored lines represent the different spatial scales considered.}
\label{fig:temp}
\end{figure}

In Figure~\ref{fig:temp} we vary the temporal scale in the $x$ axes to show the
relationship between $\mu$ and different time intervals $\Delta t$. 
We note that the speed of rank diversity increment is not increasing as in the
grammatical or spatial scales, but it has a noticeable concave shape. This
nonlinear effect is due to the fact that adding frequencies generates less
variable positions for the $N$-grams in the lists that constitute the total
time-span analyzed, therefore, $\mu$ increases until a certain time interval. Then, it starts to decrease, 
because the number of possible lists that divides the total time-span, which we
use as the denominator in the calculation of rank diversity, is lower for higher temporal scales.
Furthermore, the relation of the speed of rank diversity increment and the
temporal scale is similarly independent of the country and language. Moreover,
note how the shape of the relation between $\mu$ and time changes in some cases, considering different columns. This represents different grammatical
scales, suggesting that the grammatical scale influences on the variation of
the temporal scale.

\subsection{Spatial Scale} 

\begin{figure}
    \centering
    \includegraphics[width=0.95\linewidth,height=\textheight,keepaspectratio]{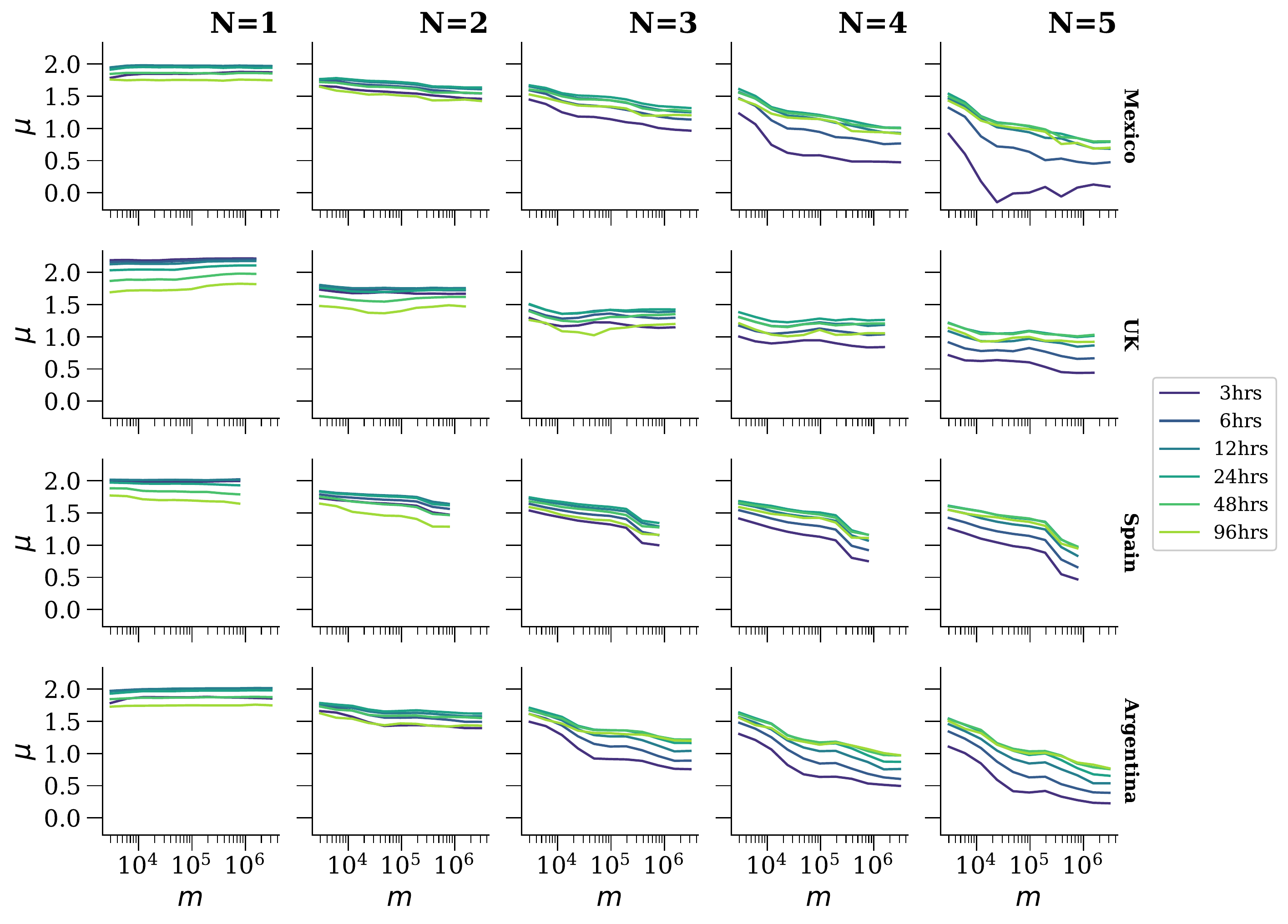}
    \captionsetup{width=.92\linewidth}
    \caption{$\mu$ estimates vs. different values of the spatial scale. The top and right labels represent the same as in
Fig.~\ref{fig:temp}. Colored lines here represent the different temporal
scales. }
    \label{fig:geograph}
\end{figure}

In each plot of Figure~\ref{fig:temp}, specially from the column $N=3$ to the
right, we can already see that given fixed grammatical and temporal scales, the
spatial scale also changes the speed of rank diversity increment. Nevertheless,
in order to see this relationship more clearly, in Figure~\ref{fig:geograph} we
plot $\mu$ versus the spacial scale. Note that for the
Spanish speaking countries, $\mu$ decreases with the spatial scale when the
grammatical scale is greater than 1, whereas,
in general, $\mu$ does not change against the spatial scale for the United
Kingdom. More detailed analyses would be required to explore potential explanations, such as whether there exists a greater homogeneity in the United Kingdom compared to the other countries, and/or whether these results reflect a difference between English and Spanish. In general $\mu$ also decreases with the grammatical scale.

\subsection{Relevance of scales} 

\begin{figure}[h!]
    \centering
\includegraphics[scale=0.60]{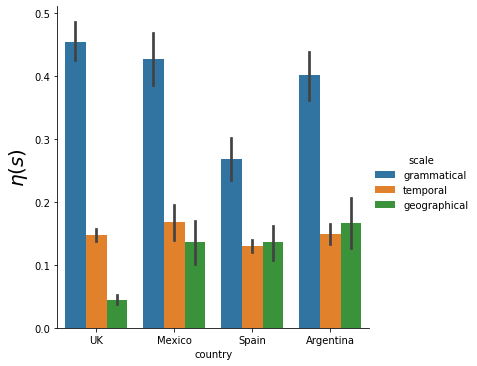}
\captionsetup{width=.92\linewidth}
    \caption{Average dispersion (2) of scale $s$ for each country and a bar chart  to compare them.}
    \label{fig:relative_importance}
\end{figure}

Now we answer the question of which scale is the most important in terms of its
effect on the variability of $\mu$ and therefore on the behavior of rank
diversity itself. To tackle this question, we measure the relative importance
of these scales using equations~\ref{eq:ETA},~\ref{eq:mean_MU},
and~\ref{eq:Dispersion}. We show in Figure~\ref{fig:relative_importance} the
results of computing the aforementioned equations for each scale and country.
It confirms that the grammatical scale accounts for the maximum dispersion
relative to the considered scales. Furthermore, the temporal and spatial scales
are both approximately equally important for all the Spanish-speaking
countries. For the United Kingdom, the spatial scale seems to have less
relevance, although more data. 

Finally, the $p$-values and the associated estimated $F$-statistic considering
the first four terms of model~\ref{eq:multiplicative} are shown in
Table~\ref{tab:Fs}.  The  low $p$-values indicate that at least one
scale is related to $\mu$, assuming that they are approximately linearly
correlated. Here we only were interested in supporting the hypothesis that changes in some
scale produce changes in the rank diversity behavior, or, in other words, that
the linear regression model provides a better fit to the data than a model with
no independent variables (with no influence of the scales that could explain
the observed variability). Specifically, we can test whether or not a
particular scale is related to $\mu$ by testing the significance of the
associated coefficient. These $p$-values are in
Table~\ref{tab:individual-pvalues}. 

\begin{table}[]
    \centering
    \begin{tabular}{|c|c|c|c|}
    \hline
     $p$-value & $X_1$ & $X_2$ & $X_3$\\
     \hline
     Mexico & $<2.0\times10^{-16}$ & $<2.0\times10^{-16}$ & $9.8\times10^{-16}$ \\
     United Kingdom & $<2.0\times10^{-16}$ &$0.125$ & $0.426$ \\
     Argentina  & $<2.0\times10^{-16}$ & $<2.0\times10^{-16}$ & $<2.0\times10^{-16}$ \\
     Spain  & $<2.0\times10^{-16}$ & $<2.0\times10^{-16}$ & $0.0183$ \\
     \hline
    \end{tabular}
    \caption{Associated $p$-values to $t$-tests of the linear coefficients. $X_1$,$X_2$, and $X_3$ represent the grammatical, spatial, and temporal scales respectively.}
    \label{tab:individual-pvalues}
\end{table}

It is worth noticing that for the United Kingdom, the temporal and spatial
scales are not so significant according to our test, compared to the grammatical scale. As previously mentioned
(Figure~\ref{fig:geograph}), the spatial scale seems practically independent
from $\mu$. However, for the temporal scale this means that a linear
approximation is not sufficient to capture the relation between these scales
and $\mu$. Thus, fitting a quadratic model is enough to find the existence of
relations for this scale, also revealing that, for this dataset, the nature of
the temporal relation with $\mu$ is non-linear, as seen in Figure~\ref{fig:temp}. 

Alternatively, to see which interaction between pairs of scales is
statistically significant, we can use the $p$-values associated to $t$-tests of
the estimated coefficients $\beta_4$, $\beta_5$, and $\beta_6$ in
model~\ref{eq:multiplicative}. Results are shown in
Table~\ref{tab:interactions}. We observe that all the interactions between the
grammatical and spatial scales and the grammatical and temporal scales are
significant, although for Argentina this is lower. Also there is a higher interaction between grammatical and spatial scales only for the United Kingdom.

\begin{table}[]
    \centering
    \begin{tabular}{|c|c|c|c|c|}
    \hline
      country & Mexico & United Kingdom & Argentina & Spain\\
     \hline
       $l-1$ &  3  &  3 &  3 & 3 \\
       $H-l$ & 326 & 296 & 326 & 266 \\
        $F$-statistic & 489.6 & 631.7 & 546 & 261\\
        $p$-value & $<2.2\times10^{-16}$ & $<2.2\times10^{-16}$ &  $<2.2\times10^{-16}$ & $<2.2\times10^{-16}$\\
        \hline
    \end{tabular}
    \caption{p-values associated to the $F$-statistic, which represents the significance of the regression, \emph{i.e.}, that at least one scale is contributing to explain the changes in rank diversity. $l$ is the number of parameters used and $H$ the number of observations.}
    \label{tab:Fs}
\end{table}

\begin{table}[]
    \centering
    \begin{tabular}{|c|c|c|c|}
    \hline
     $p$-value & $X_1$*$X_2$ & $X_1$ * $X_3$ & $X_2$ * $X_3$\\
     \hline
     Mexico & $<2.2\times10^{-16}$ & $<2.2\times10^{-16}$ & $0.57$ \\
     United Kingdom & $<8.2\times10^{-5}$ &$<2.2\times10^{-16}$ & $0.12$ \\
     Argentina  & $<2.2\times10^{-16}$ & $<2.2\times10^{-16}$ & $<2.6\times10^{-4}$ \\
     Spain  & $<2.2\times10^{-16}$ & $<2.2\times10^{-16}$ & $0.47$ \\
     \hline
    \end{tabular}
    \caption{Associated $p$-values to $t$-tests of the interaction coefficients. $X_1$,$X_2$, and $X_3$ represent the grammatical, spatial, and temporal scales respectively.}
    \label{tab:interactions}
\end{table}
\subsection{Special tokens} 

In this section we focus on analyzing special tokens commonly used in Twitter:
emojis, hashtags and mentions. We studied the most frequent occurrences within
each country and the rank diversity for Argentina, Mexico, Spain and the
United Kingdom. 

An \emph{emoji} is a pictogram, logogram, ideogram, or smiley used in
electronic messages and web pages. The primary function of emojis is to fill in
emotional cues otherwise missing from typed conversation and refers to pictures
that can be represented as encoded characters.  Emojis have become widely used
to communicate emotion. It is useful to emphasize our communication with
body language and facial expressions, which are lacking in texts. Thus, they can be
complemented with emojis.

\begin{figure}
 \centering 
 \includegraphics[scale=0.5]{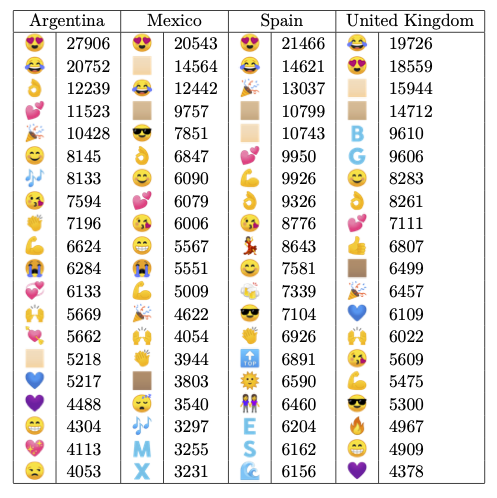}
 \captionsetup{width=.92\linewidth}
\caption{Comparison of the most frequent emojis and modifiers shown in  Argentina, Mexico,
Spain and the United Kingdom in geolocalized tweets from  the year 2014.}
\label{fig:emojis}
\end{figure}

Figure~\ref{fig:emojis} can be read from top to bottom, showing the most
frequent emojis in the first rows from the top\footnote{Devanagari characters (popular
in the UK) were removed as these are not emojis.}.  We can see \textit{Smiling
Face with Heart-Shaped Eyes} and \textit{Face with Tears of Joy}  as the most
commonly expressed emotions, followed by other expressions such as happiness,
hearts and strength. It is interesting to notice that some emojis are popular
across all countries, but also there are some that are prevalent only in some
countries. For example, the \textit{Woman Dancing} and \textit{Sun with Face}
emojis are popular only in Spain, \textit{Sleeping Face} only in Mexico, \textit{Unamused Face} only in
Argentina, while \textit{Fire} appears only in the UK.

We can notice that within the most frequent unicode emoji symbols are some \textit{Emoji
Modifier Fitzpatrick}. Emoji characters can be modified to use one of five
different skin tone modifiers. Each tone is based on the Fitzpatrick Scale. The
Fitzpatrick scale is a numerical classification schema for human skin color
\cite{fitzpatrick1988validity} developed in 1975 by American dermatologist
Thomas B. Fitzpatrick. Argentina only has in its top list \textit{Light Skin
Tone}. Spain also includes \textit{Medium-Light Skin Tone}. Mexico and the UK
have \textit{Medium Skin Tone} as well. \textit{Medium-Dark Skin Tone} and
\textit{Dark Skin Tone} do not appear in any of these lists.


A \emph{hashtag} is a metadata tag that is prefaced by the hash symbol,
\textit{\#}. Hashtags are used on platforms such as Twitter and Instagram as a
form of user-generated tagging that enables cross-referencing of content, that
is, sharing a topic or theme. They are useful for finding content of similar
interest. 

It is important to note that hashtags are neither registered nor controlled by
any one user or group of users. They do not contain any set definitions,
meaning that a single hashtag can be used for any number of purposes, and that
the accepted meaning of a hashtag can change with time.

\begin{table}[h] 
\centering
\tiny
    \begin{adjustbox}{width=350pt}
\begin{tabular}{|l|l|l|l|l|l|l|l|}
\hline
  \multicolumn{2}{|c|}{Argentina} & \multicolumn{2}{c|}{Mexico} & \multicolumn{2}{c|}{Spain} & \multicolumn{2}{c|}{United Kingdom} \\
  \hline
  
 \#trndnl  & 9518 &  \#trndnl  & 10836 & \#trndnl  & 19328 &  \#nowplaying   & 52085 \\
\#buenosaires  & 5022 & \#cdmx    & 6951 &  \#madrid  & 14552 & \#london  & 35934 \\
\#argentina      & 2372 & \#mexico  & 6540 & \#barcelona & 14167 & \#job    & 25486 \\
\#me   & 1646 & \#mexicocity & 3631 &  \#incdgt & 13367 & \#tnc     & 24897 \\
\#cordoba    & 1502 & \#job & 3508 & \#dgt	& 12978 & \#trndnl &	24319 \\
\#rosario &  1419 & \#hiring  & 3135 &  \#meteocat   & 12260 & \#areacode & 23084 \\
\#love    & 1269 &  \#monterrey  & 2870 & \#endomondo   & 9689 & \#hiring &	22587 \\
\#selfie  & 1088 &  \#endomondo & 2626 & \#meteo	& 8736 &  \#tides     & 20236 \\
\#friends &  1069 & \#guadalajara   & 2433 & \#spain	& 8649 & \#ktt	& 19237 \\
\#job   & 961 &  \#endorphins  & 2299 &  \#blanco &	8644 & \#weather    & 12024 \\
\#endomondo & 939 & \#friends   & 2049 &  \#endorphins &	8559 &  \#careerarc & 11841 \\
\#viernes  & 911 &  \#méxico  & 1860 &  \#324meteo &	8468 & \#essex &	11016 \\
\#amigos  & 891 &  \#love  & 1758 &  \#meteo   & 3127 &  \#broadbandcompareuk &	10881 \\
\#repost  & 873 &  \#careerarc  & 1575 &  \#retención & 8189 & \#bestbroadband & 10641 \\
\#hiring    & 868 & \#photo  & 1486 &  \#precaución &	7798 &  \#photo  & 10492 \\
\#night &  807 & \#jobs  & 1370 & \#elcatllar   & 6844 & \#ukweather	& 9978 \\
\#endorphins   & 804 & \#quieremeamame & 1333 &  \#obra  & 5398 & \#endomondo &	9770 \\
\#domingo  & 774 &  \#puebla  & 1215 &  \#arameteo    & 4910 &  \#jobs &	9375 \\
\#sabado & 727 & \#selfie   & 1183 & \#sevilla  & 4890 &  \#endorphins  & 8988 \\
\#carlosrivera &  719 & \#travel & 1069 &  \#amarillo   & 4030  & \#stalbans & 7498 \\

  \hline
\end{tabular}
\end{adjustbox}
\caption{Most frequent hashtags in Argentina, Mexico, Spain and United Kingdom in 2014 geolocalized tweets.}
\label{tab:hashtags}
\end{table} 

Table~\ref{tab:hashtags} shows that \textit{\#trndnl} is the most common
hashtag, which is related to popularity and trending topics. Also important
cities of each country are mentioned: Buenos Aires, Cordoba, and Rosario in
Argentina; CDMX/Mexico City, Monterrey, Guadalajara, and Puebla in Mexico;
Madrid, Barcelona, and Sevilla in Spain; and London and St Albans in the UK. It
is interesting that only in Europe weather-related hashtags are popular.

A \emph{mention} is a tweet that contains another person’s username anywhere in
the body of the tweet. 
User mentions are identified with the $@$ symbol within tweets.

\begin{table}[h]
\centering
\footnotesize
\tiny
    \begin{adjustbox}{width=350pt}
\begin{tabular}{|l|l|l|l|l|l|l|l|}
\hline
  \multicolumn{2}{|c|}{Argentina} & \multicolumn{2}{c|}{Mexico} & \multicolumn{2}{c|}{Spain} & \multicolumn{2}{c|}{United Kingdom} \\
  \hline
  
@clubsolotu\_arg  & 724 & @aicm3 & 3965 & @canalfiesta     & 1124 & @nationalrailenq & 4797 \\

@vale975         & 649 & @cinemex         & 2629 & @aena  & 1068 & @heathrowairport & 2088 \\

@pabloalucero    & 543 & @cinepolis       & 2352 & @aenaaeropuertos & 842 & @starbucksuk     & 1480 \\

@rkartista & 432 & @germanmontero5  & 2296 & @dominguezja &	830 & @luvthenorth444  & 1401 \\

@aa2000oficial   & 417  & @grupointocable & 2122 & @oficialmaki     & 671 & @brewdog         & 1030 \\

@mauriciomacri   & 368 & @sonadoraeterno  & 2108  & @oficiallamorena & 570 & @simmons2k       & 928  \\

@abrahammateomus & 339 & @mariobautista\_  &	1878 & @adif\_es         & 462 & @lynnie26blue    & 903 \\

@lucianopereyra  & 325  & @smartfit\_mex    & 1755 & @willylevy29     & 395 & @costacoffee &	902\\

@todonoticias & 263 & @walmartmexico   & 1274 & @realmadrid      & 321 & @lizbussey       & 863 \\

@c5n    & 243 & @galeriasmx   & 1010 & @pinedademar  & 297  & @babs200475  & 819 \\

@infobae         & 179  & @chilismexico    & 789 & @adry60go        & 288 & @shelleym1974    & 774 \\

@radiomitre      & 169 & @aeropuertodemty & 729 & @pablo\_iglesias\_ & 278 & @shelleym1974    & 774 \\

@rialjorge       & 163 & @aeropuertosgap  & 695  & @sanchezcastejon	& 273 & @luvyorkshire444 & 	693\\

@cfkargentina    & 156 & @solosanborns & 673 & @psoe &	265 & @westendlanegirl	& 659\\

@marialuizateodo & 153 & @banamex  & 654  & @fcbarcelona &	242 & @feelingpeacenmw &	634 \\

@starbucksar & 153 & @lacasadetono & 644 & @barcelona\_cat &	234 & @harrods	& 625\\

@niallofficial & 152 & @oasis\_coyoacan  & 625 & @ahorapodemos & 213 & @kathb24 & 576 \\

@sole\_pastorutti & 152  & @auditoriomx & 594  & @renfe	& 202 & @selfridges	& 560\\

@brigitte2300 &	144 & @perisur	& 566 & @el\_pais & 199 & @visitlondon	& 481 \\

@lanacion &	144 & @tuado  & 559 & @mariajosesernas & 190 & @skynews         & 466 \\

 \hline
\end{tabular}
\end{adjustbox}
\caption{The most frequent users mentioned in Argentina, Mexico, Spain and the United Kingdom in geolocalized tweets from the year 2014.}
\label{tab:mentions}
\end{table} 

Table~\ref{tab:mentions} shows the most common user mentions. They highlight
meeting places such as shopping centers, cinemas, and airports; and famous
companies or people (politicians, artists, sportspeople). Differences can be
seen per country, suggesting variations in the usage of Twitter at the time.
For example, Argentina has several mentions of artists, Mexico has
several mentions of commercial franchises, Spain includes soccer teams and
political parties, and the most mentioned account in the UK by far is that of
the National Rail network.

It is important to notice that these mentions are from geolocated tweets, which
are only a small fraction of all tweets. Thus, these mentions might be biased
and the most popular accounts might vary from those listed here.

The rank diversity curves of emojis, hashtags, and user mentions can be
approximated with a sigmoid curve, as with many other
phenomena~\cite{Iniguez2021}.  In all cases, user mentions are the most diverse
feature of the country and the emojis are the least diverse feature.

\begin{figure*}[t] 
\centering
  \begin{tabular}{cc}
  \includegraphics[scale=0.35]{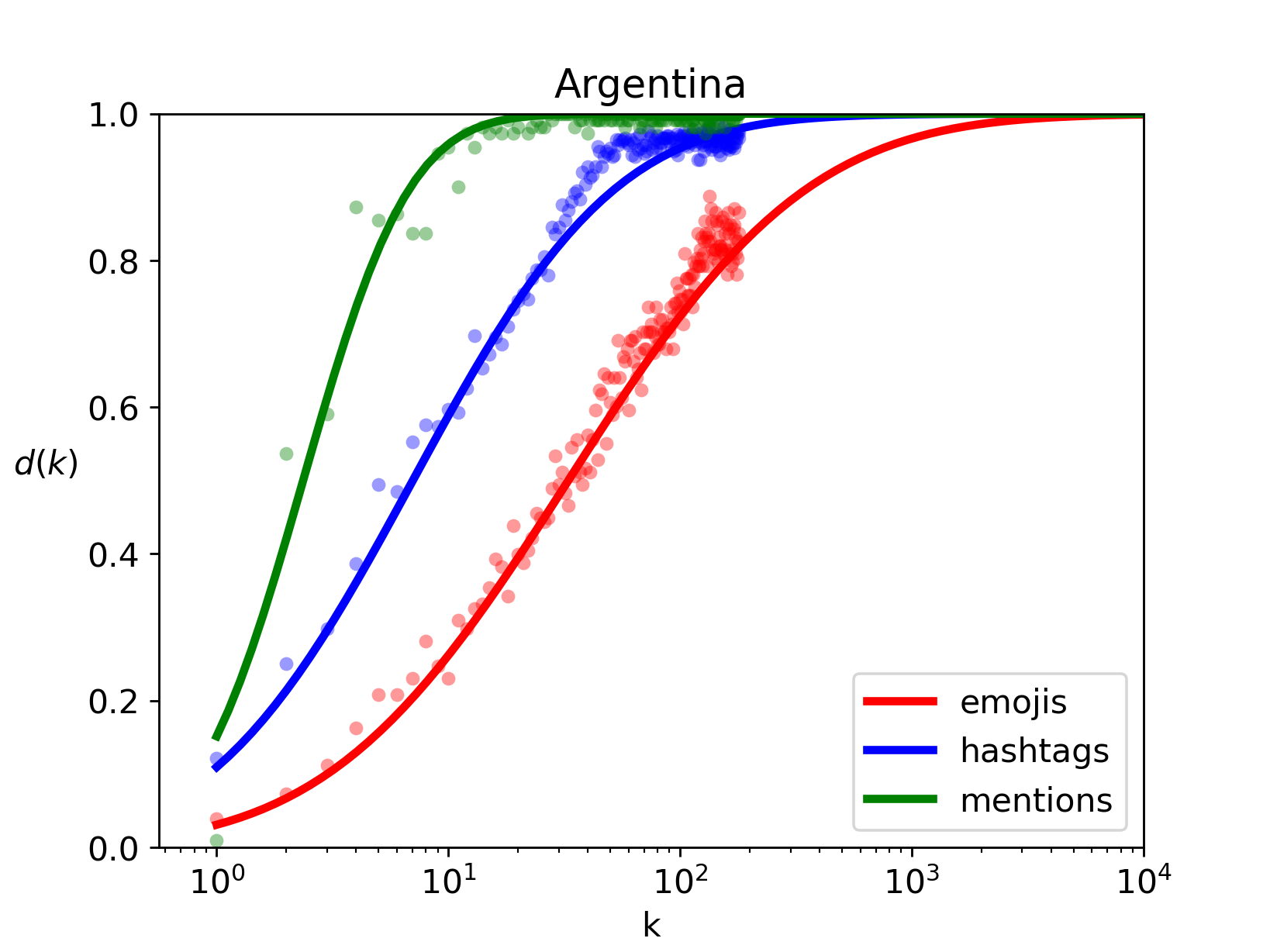} & 
  \includegraphics[scale=0.35]{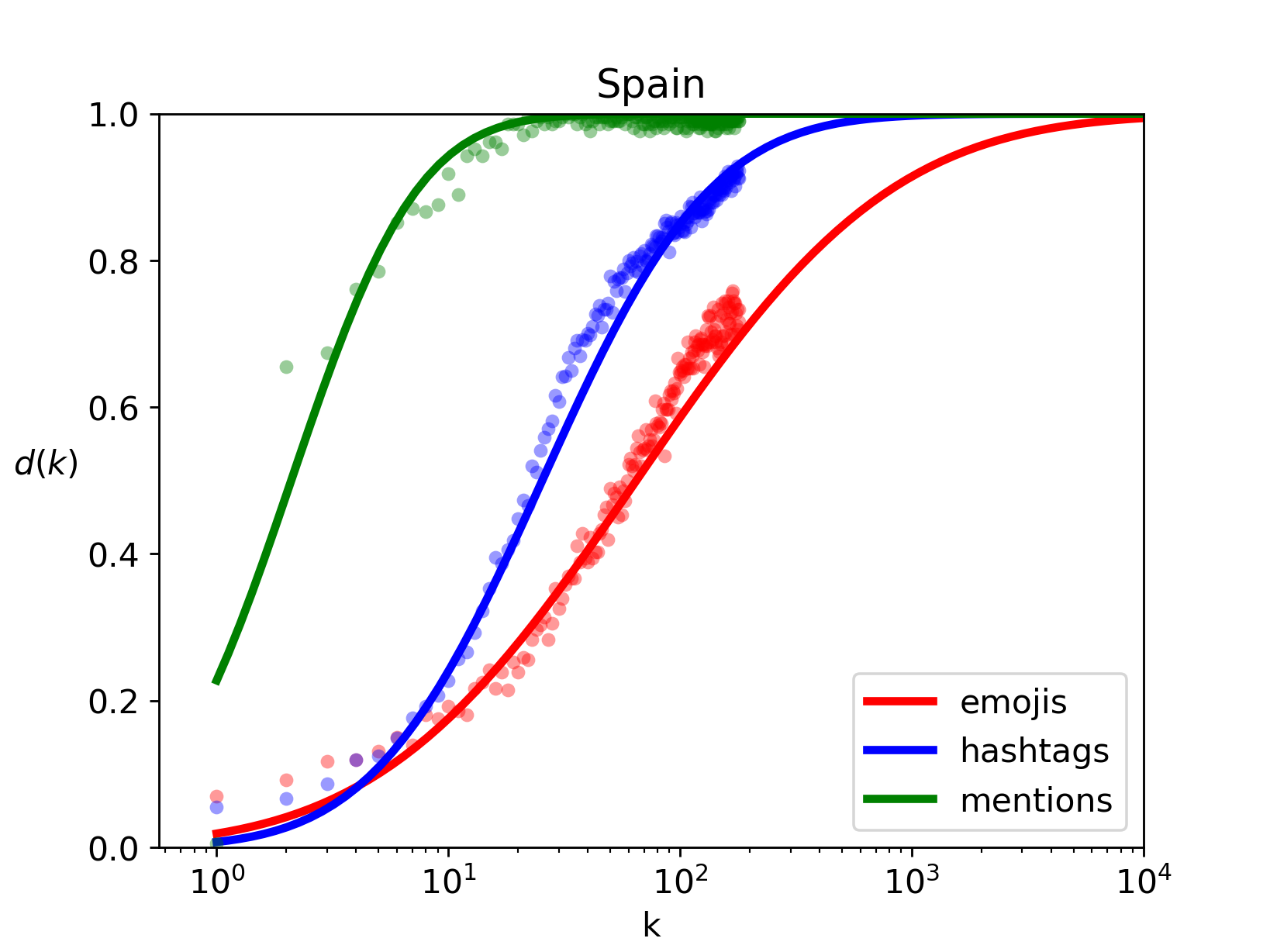} \\
  \includegraphics[scale=0.35]{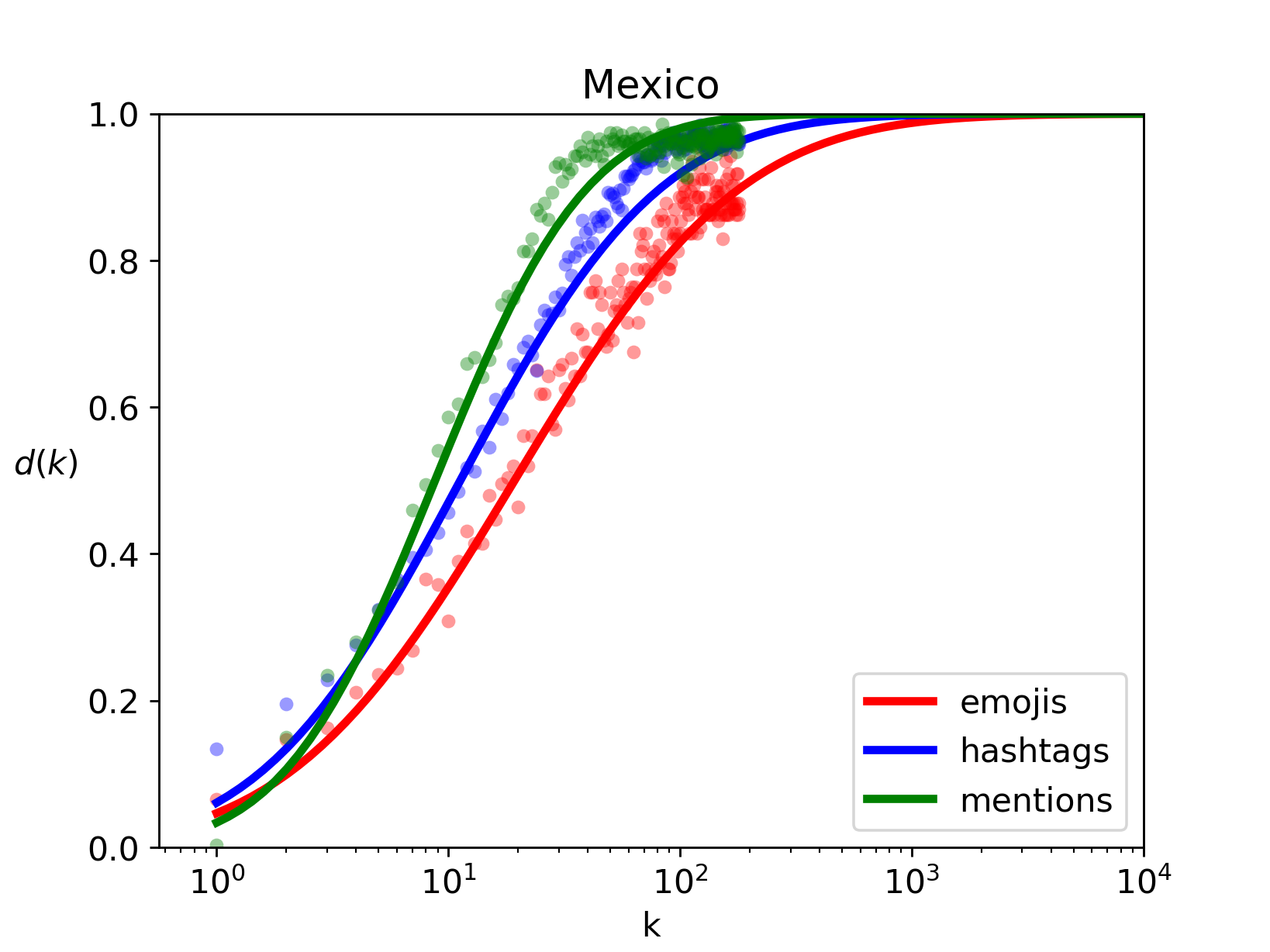} &
   \includegraphics[scale=0.35]{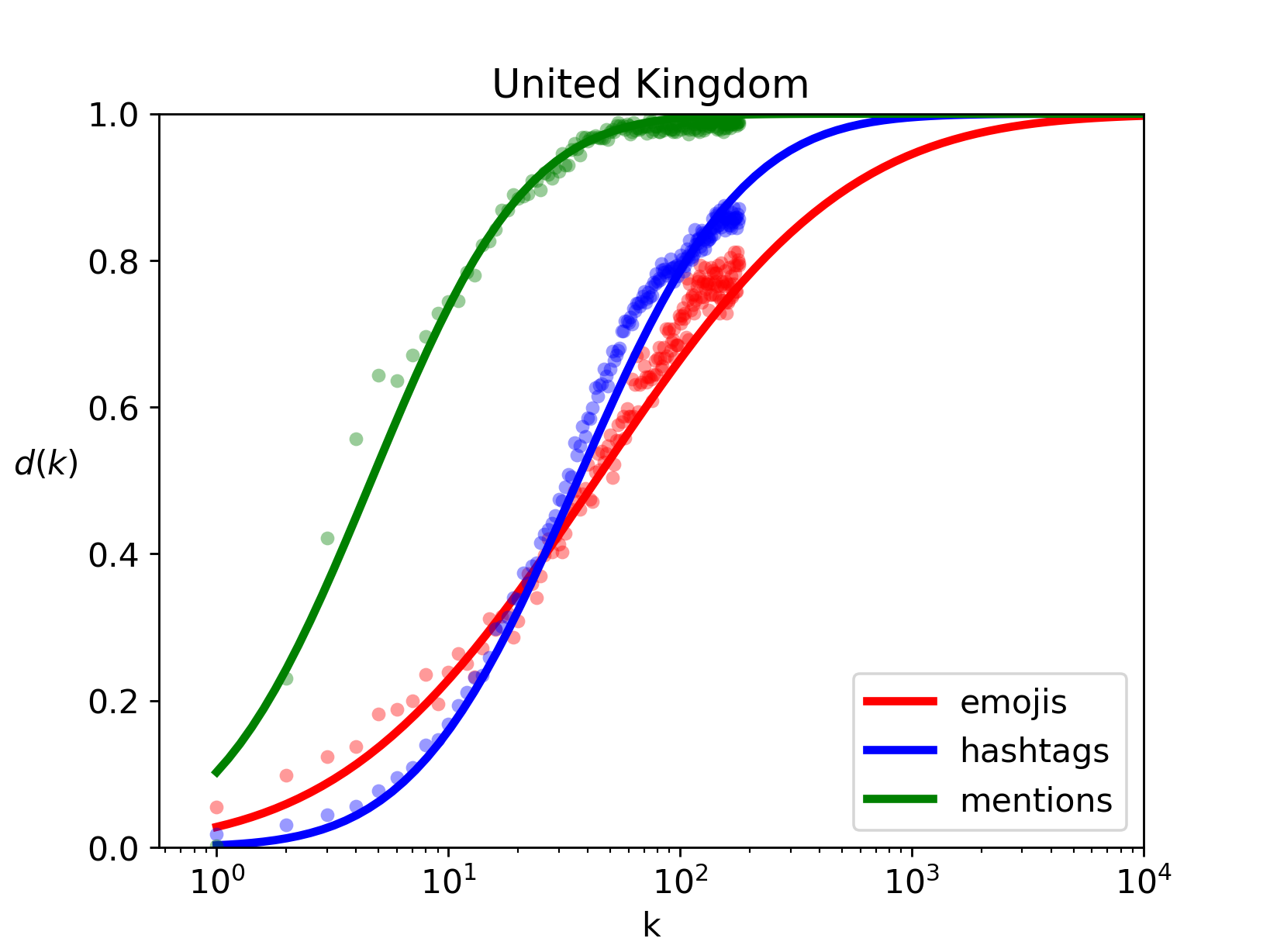}
 \end{tabular}
 \captionsetup{width=.92\linewidth}
\caption{Rank diversity $d(k)$ of emojis, hashtags and mentions in Argentina, Mexico, Spain and United Kingdom in a time interval $\Delta t$ of $24$ hours.}
\label{fig:rank}
\end{figure*} 
\section{Conclusions} 




Analyzing geolocalized Twitter data, we were able to study language use at different spatial, temporal, and grammatical scales.
Which scales are more relevant for languages? All of them, but the grammatical
is more relevant than the spatial and temporal, at least for the statistical measures considered here.
This suggests that the relation
of the considered scales with rank diversity cannot be completely understood in
detail using just one of the scales, even when the grammatical scale is the
most strongly related to changes in the speed of rank diversity increment. 



It is interesting that the sigmoid curve correctly describes rank diversity
curves in all the considered combinations of spatial and temporal scales. Thus,
not only this function adequately fits rank diversity curves for different
languages and different $N$-grams as our previous research showed for temporal
scales of years, but it also correctly fit rank diversity curves in the smaller
temporal scales considered here and geographical regions. This suggests that
the shape of the rank diversity curve is derived from mechanisms that are not
affected by changes in language or scales~\cite{Iniguez2021}.
Also, notice that the diversity of monograms are not affected by the spatial
scale. However, as the grammatical scale increases, the rank diversity curves
do change with the spatial scale. This is probably because word usage across
regions should be more similar than phrase usage. In other words, language use
variability increases at higher grammatical scales.

The evidence of interactions between scales means that rank diversity exhibits
different within-scale behavior depending on which values the remaining scales
adopt. This effect is clearly noticeable between temporal and grammatical
scales and between spatial and grammatical scales. Nonetheless, our results do
not support that an interaction exists between the spatial and temporal scales.
In general, the speed of increment of rank diversity is greater at higher
scales in both grammatical and geographical cases for the Spanish-speaking
countries. For example, in terms of 2-grams, this means that the rate of
increment of the number of different 2-grams that appear in the ranks during a
time span of one year divided into periods of $t$ hours, where $t$ is one of
the possible temporal scale values, increments as the spatial scale increases.

We compared the most frequent emojis in different languages and countries,
suggesting that emojis are nonverbal symbols that reflect cultural differences
between Twitter users and their geographical locations.  They can also reflect
the collective sentiment~\cite{Bollen:2011} of each country and perhaps even
certain biases.  Hashtags are metadata embedded in a social network, Twitter in
this case. The social aspect of it is in the ability to create communities but
also evokes emotions and express feelings. Moreover, hashtags can help
recognize the relevant topics and events of a community. Thus, understanding
their dynamics can lead to several potential insights.  Mentions in Twitter are
used to refer to another user account. It shows the most popular business,
people or accounts. Their change in time and space reflects how their relevance
varies.  The use of all of these special tokens also can be approximated with a
sigmoid curve in their rank diversity. For different countries, emojis seem the
most stable, then hashtags, and mentions are the most volatile.

During the COVID-19 pandemic, the sharing of misinformation on social media has
become a major focus in academic studies. For example, Pennycook, \emph{et
al.}~\cite{Pennycook2021} found that shifting attention to accuracy can reduce
misinformation online. An interesting extension of this work would be to study
the statistical linguistics of misinformation on Twitter.

\begin{backmatter}

\section*{\large{Declarations}}

\section*{Availability of data and materials}
Data are upon request from the authors.

\section*{Competing interests}
  The authors declare that they have no competing interests.

\section*{Funding}
This work was supported by UNAM's PAPIIT IN107919, IG101421 and IV100120, and 
CONACyT 285754 grants.

\section*{Author's contributions}
AMG, CG and CP designed research; FSP, RLA, DPM, PR and EC analyzed the data and performed numerical simulations; CG and CP analyzed
results and wrote the paper, AMG provided the dataset. All authors read and approved the final manuscript.

\bibliographystyle{bmc-mathphys}
\bibliography{carlos,refs}

\end{backmatter}
\end{document}